# MCANet: A Multi-Scale Class-Specific Attention Network for Multi-Label Post-Hurricane Damage Assessment Using UAV Imagery

Zhangding Liu,[1] Neda Mohammadi,[2] and John E. Taylor[3*]

[1]Ph.D. Student, School of Computational Science and Engineering, Georgia Institute of Technology, 790 Atlantic Dr NW, Atlanta, GA 30332, United States; e-mail: zliu952@gatech.edu

[2]Sydney Horizon Fellow, The University of Sydney, NSW 2006, Australia; e-mail: neda.mohammadi@sydney.edu.au

[2]School of Civil and Environmental Engineering, Georgia Institute of Technology, 790 Atlantic Dr NW, Atlanta, GA, 30332, United States; e-mail: nedam@gatech.edu

[3*]Professor, School of Civil and Environmental Engineering, Georgia Institute of Technology, 790 Atlantic Dr NW, Atlanta, GA, 30332, United States; e-mail: jet@gatech.edu (corresponding author)

## ABSTRACT

Hurricanes cause widespread damage to buildings, roads, and other infrastructure, making timely post-disaster damage assessment critical for emergency response and recovery planning. Unmanned aerial vehicle (UAV) imagery provides high-resolution observations of affected areas, but post-hurricane scenes are difficult to classify because multiple damage categories often co-occur within the same image, appear at different spatial scales, and include visually similar severity levels as well as rare but operationally important classes. To address these challenges, this study presents MCANet, a multi-label classification framework for post-hurricane UAV damage assessment. MCANet integrates a Res2Net-based backbone for multi-scale representation with class-specific residual attention to aggregate spatial evidence separately for each damage category. Evaluation on the RescueNet dataset, which includes 4,494 UAV images collected after Hurricane Michael and annotated with 10 damage categories, shows that MCANet achieves the highest mean average precision (mAP) among the evaluated models, with an mAP of 91.37%. Compared with Vision Transformer (ViT-B/16), MCANet improves mAP by 1.90 percentage points while using approximately 50% fewer parameters and 52% fewer giga floating-point operations (GFLOPs). Class-specific residual attention contributed an mAP increase of 1.52 percentage points over class-agnostic attention (CA); the Res2Net backbone alone produced concentrated gains on scale-heterogeneous or underrepresented classes such as Road Blocked and Pool. The single-head configuration achieved the highest overall mAP, whereas the two-head configuration provided targeted gains for Road Blocked and Building Major Damage. MCANet supports rapid post-disaster image screening and provides structured outputs for downstream emergency-management workflows.

## Introduction

Hurricanes are among the most destructive natural disasters, severely impacting communities across the United States by causing widespread infrastructure damage, service disruption, population displacement, and significant economic losses (Balaguru et al. 2023; Grinsted et al. 2019; Klotzbach et al. 2018; Young and Hsiang 2024). From 1900 to 2017, the United States was impacted by 197 hurricanes, resulting in 206 landfalls. When adjusted to 2018 economic conditions, the total damage from these events amounts to nearly $2 trillion, averaging close to $17 billion in losses per year (Weinkle et al. 2018). Beyond economic losses, hurricanes also disrupt human mobility, with studies showing short-distance trips increasing and long-distance travel sharply decreasing during disasters (Wang and Taylor 2014, 2016). As climate change progresses and coastal populations grow, hurricane-related losses are expected to continue to increase (Emanuel 2005; Jing et al. 2024). These trends indicate the need for tools that can automatically classify and assess post-disaster infrastructure damage to support emergency response and minimize losses.

Timely and accurate classification of post-hurricane damage is vital for coordinating emergency response efforts, prioritizing the allocation of resources, and facilitating rapid recovery (Al Shafian and Hu 2024; Kaur et al. 2021; Sun and Huang 2023). Quick identification of the most severely impacted areas can guide rescue operations, ensuring that aid reaches the most vulnerable populations first (Chou et al. 2017). Additionally, a detailed understanding of the extent and type of damage is essential for infrastructure repair and reconstruction planning, enabling communities to restore essential services and return to normalcy as quickly as possible (Liu et al. 2022).

Current post-disaster damage assessment methods face several limitations that hinder timely and accurate decision-making. Ground-based surveys are often slow, labor-intensive, and geographically constrained, and they expose rescue teams to potential hazards (Aela et al. 2024; Einizinab et al. 2023; Ejaz and Choudhury 2024; Spencer et al. 2019). Satellite imagery provides broad regional coverage and is frequently used for large-scale assessments (Al Shafian and Hu 2024), but it suffers from low spatial resolution, reliance on clear weather, and limited viewing angles. Social media has emerged as a supplementary data source offering near real-time observations, but it is highly user-dependent, often lacks geotagging accuracy, and varies greatly in content quality (Salley et al. 2024). In contrast, an Unmanned aerial vehicle (UAV) provides high-resolution, near real-time imagery with the ability to capture building and road level damage at flexible altitudes and angles, making it a promising tool for rapid and reliable post-disaster damage assessment.

Once high-quality imagery is collected, the next challenge is to accurately classify the observed damage. Unlike conventional image classification datasets such as ImageNet (Deng et al. 2009), where each image typically contains a single, well-defined object, post-disaster aerial imagery is more complex. A single image may exhibit multiple, coexisting damage types—such as collapsed buildings, road blockages, and flooded areas—each occupying different spatial regions and varying greatly in scale and visibility. Large-scale destruction may span the entire image, while smaller features like debris are confined to localized areas and often obscured by clutter. Despite this complexity, many existing models adopt a single-label multi-class classification approach, assigning only one dominant damage category to an image or patch (Cao and Choe 2020; Cheng et al. 2021; McCarthy et al. 2020). This oversimplification mirrors earlier practices in remote sensing, where traditional scene classification methods relied on representing each image with the most prominent terrain feature (Qi et al. 2020). High-resolution UAV imagery often contains multiple semantically rich object types within the same scene, rendering single-label assumptions inadequate for disaster contexts (Tseng 2023). As a result, single-label models tend to oversimplify real-world disaster scenes (Shao et al. 2018), where overlapping or interdependent damage types are common, and reduce model effectiveness for response planning. Although object detection and semantic segmentation models can support multi-label outputs, they require dense, pixel-level annotations that are labor-intensive and time-consuming—up to one hour per image in the RescueNet dataset (Rahnemoonfar et al. 2023; Sirhan et al. 2024). This high annotation cost hinders scalability in time-critical disaster scenarios. In contrast, multi-label classification offers a more practical solution: it relies on coarse image-level labels while still enabling recognition of multiple damage types within the same image.

Despite recent advances in multi-label learning for aerial imagery (Pitakaso et al. 2024; Stoimchev et al. 2023; Sumbul et al. 2019; Tseng 2023), existing methods remain limited for post-hurricane UAV damage assessment in three ways. First, visual damage cues span a wide spatial range, from localized objects to scene-wide patterns. Second, multiple damage categories may co-occur in different regions of the same image, requiring the model to distinguish class-specific spatial evidence rather than relying only on shared global representations (Kopiika et al. 2025; Zhou et al. 2023). Third, visually adjacent damage severity levels can be difficult to separate, while low-frequency but operationally important categories such as Road Blocked remain challenging to identify consistently (Liu et al. 2025a; Sumbul and Demir 2020). These limitations are consequential for disaster response because incomplete or ambiguous image-level interpretation can affect downstream damage mapping, road accessibility assessment, and prioritization of emergency resources.

These challenges motivate MCANet, a multi-scale class-specific attention network for multi-label post-hurricane damage classification from UAV imagery. The main contributions of this study are threefold. First, UAV-based post-hurricane damage assessment is formulated as an image-level multi-label classification task, allowing a single image to represent co-occurring infrastructure, road, water, vehicle, and vegetation. This formulation supports rapid scene-level damage screening and complements segmentation-based approaches when pixel-level localization is not required. Second, MCANet integrates a Res2Net-based multi-scale backbone with class-specific residual attention (CSRA) to address scale variation and category-dependent spatial evidence in post-hurricane scenes. A multi-head attention extension is further evaluated to examine whether complementary attention sharpness benefits difficult categories. Third, the framework is systematically evaluated on the RescueNet dataset through baseline comparison, per-class analysis, component ablation, multi-head analysis, class activation map visualization, and computational benchmarking. The evaluation examines not only overall classification accuracy but also how the proposed components affect difficult categories.

## Literature Review

### Deep Learning Approaches for Post-Hurricane Damage Classification

Recent advancements in deep learning have enhanced automated damage classification of civil infrastructure (Tong et al. 2025). Early studies primarily focused on binary classification, typically distinguishing between "damaged" and "undamaged" structures. McCarthy et al. (2020) used multispectral satellite imagery to classify healthy versus degraded vegetation after Hurricane Irma. Similarly, Cao and Choe (2020) employed a convolutional neural network (CNN) to distinguish flooded or damaged buildings from undamaged ones following Hurricane Harvey, achieving 97.08% accuracy. While effective, their study also noted key limitations of satellite imagery, including low spatial resolution and inconsistent label quality.

To support more detailed damage interpretation, researchers have advanced from binary to multi-class classification frameworks. Li et al. (2018) developed a semi-supervised approach combining Simple Linear Iterative Clustering segmentation, convolutional autoencoders, and a fine-tuned CNN to classify aerial image patches into five damage levels. Tested on post-Hurricane Sandy imagery, their method achieved an accuracy of 88.3%, indicating reliable performance with limited labeled data. Cheng et al. (2021) proposed a two-stage stacked preliminary damage assessment architecture using UAV footage from Hurricane Dorian. Their model classified building damage into six ordered levels (Level 0: no damage to Level 5: destroyed or under construction), reaching 61% accuracy and 90% within ±1 level on seen data. Berezina and Liu (2022) implemented a coupled CNN with U-Net-based segmentation and ResNet classification on the xBD satellite dataset (Hurricane Michael), targeting four damage categories: undamaged, minor, major, and destroyed. Their model addressed class imbalance with focal loss and achieved 86.3% overall accuracy, outperforming traditional baselines while highlighting the difficulty of distinguishing visually similar classes.

Beyond classification, Pi et al. (2021) applied semantic segmentation and object detection to the Volan2019 UAV dataset, which includes pixel-level labels for nine disaster-related categories. Using Mask Region-based Convolutional Neural Network (Mask R-CNN) and Pyramid Scene Parsing Network (PSPNet), they achieved a mAP of 51.54% and a mean intersection over union (IoU) of 32.17%. While their approach enables detailed location recognition, it requires costly pixel-level annotations. In this context, multi-label classification provides a practical alternative, requiring only image-level labels while still capturing multiple coexisting damage types within a scene.

## Multi-Label Classification in Disaster Damage Assessment

Multi-label classification has emerged as an effective paradigm for analyzing complex visual scenes, particularly in remote sensing and disaster response. Unlike single-label models, multi-label approaches can capture the co-occurrence of diverse categories within a single image, making them well-suited for tasks involving aerial or satellite imagery. This formulation has demonstrated superior descriptive capabilities (Stoimchev et al. 2023; Tseng 2023) and has been adopted in applications such as land cover mapping (Sumbul and Demir 2020; You et al. 2023), image retrieval (Shao et al. 2020), and post-disaster data curation (Ho Ro et al. 2024; Pitakaso et al. 2024). The BigEarthNet benchmark (Sumbul et al. 2019) further highlights the need for multi-label learning in remote sensing due to the heterogeneous nature of overhead images. In disaster informatics, Park et al. (2022) demonstrated the advantages of a multi-output CNN for post-earthquake image tagging, showing that unified multi-label models can improve classification efficiency and scalability in large reconnaissance datasets.

Post-hurricane aerial images remain difficult to analyze because multiple damage types, such as road blockages, building collapses, debris, and flooding, often appear together within the same scene and at different spatial scales. These complex scenes often exhibit strong intra-class variation and inter-class similarity, which complicates the task of assigning accurate labels (Ro and Gong 2024). Moreover, damage may range from large-area inundation patterns to fine-grained structural details, requiring models that can capture both global context and local evidence. Conventional models often fall short in this setting, as they are typically limited by fixed receptive fields and lack mechanisms to effectively aggregate multi-scale context (Kopiika et al. 2025; Liu et al. 2025b). Another challenge is that cluttered backgrounds and visual ambiguity can reduce discriminative ability, leading to confusion among overlapping or visually similar damage categories. As noted by Liu et al. (2025a), standard CNNs often struggle to differentiate visually similar classes without architectural enhancements, which can lead to misclassification of low-frequency but operationally important damage categories in imbalanced disaster datasets. These limitations motivate methods that can learn multi-scale representations while preserving class-specific discriminative information.

Recent research has attempted to address parts of these challenges through attention mechanisms and multi-scale feature fusion. For example, You et al. (2023) introduced an attention-enhanced DenseNet using spatial and channel attention to better highlight relevant features. Kaur et al. (2023) proposed a hierarchical transformer for building damage classification, fusing features across multiple resolutions and achieving strong results on the xBD benchmark. These studies highlight the promise of attention-aware, multi-scale design for disaster assessment. However, existing methods typically emphasize either multi-scale feature learning or attention-based enhancement, rather than jointly addressing both within a unified multi-label framework for post-hurricane UAV imagery. In particular, limited attention has been given to class-specific spatial discrimination in scenes where multiple damage categories co-occur. MCANet is therefore developed to combine multi-scale representation with class-specific attention for more precise multi-label damage assessment.

## Methods

MCANet is developed around three characteristics of post-hurricane UAV imagery: scale-varying damage cues, co-occurring categories in different image regions, and visually similar severity levels requiring category-specific spatial discrimination. The following subsections describe the flowchart of MCANet, multi-scale backbone, class-specific residual attention module, and multi-label training loss.

### Flowchart of the Proposed Approach

The proposed MCANet framework converts a post-hurricane UAV image into a structured multi-label prediction vector, where each output dimension represents the predicted presence of one damage-related category. As shown in Fig. 1, the framework consists of three main components: a Res2Net-based multi-scale feature extraction backbone, a class-specific residual attention module, and a sigmoid-based multi-label prediction layer.

Given an input UAV image, the Res2Net backbone first extracts a multi-scale feature representation. This backbone is chosen to capture visual cues with substantially different spatial extents in post-hurricane scenes. The extracted feature representation is then passed to the CSRA module, which aggregates spatial evidence separately for each damage category. The class-specific design is needed because different labels in the same image may correspond to different spatial regions.

The attention-enhanced class representations are used to produce one logit for each damage category. These logits are transformed with sigmoid activation to obtain independent class probabilities, allowing multiple labels to be assigned to the same image. The primary MCANet configuration uses a single CSRA head. A multi-head extension with different attention temperature settings is also evaluated to examine whether complementary attention sharpness improves selected difficult categories. The detailed formulation of the feature extraction module, attention module, and training loss is provided in the following subsections.

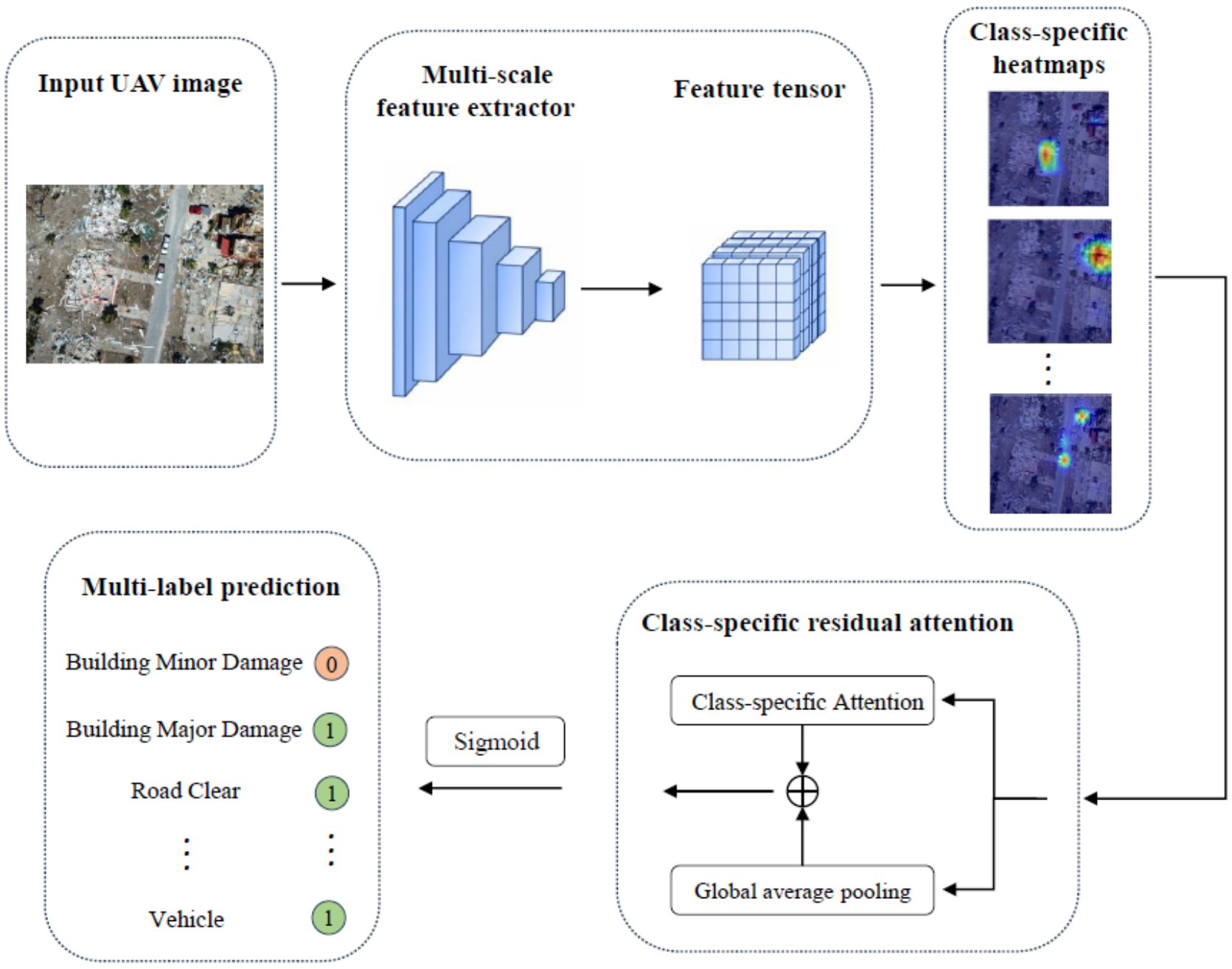

**Fig. 1.** Overview of the proposed MCANet framework for multi-label post-hurricane damage classification. The framework extracts multi-scale features using a Res2Net backbone, aggregates category-specific spatial evidence through class-specific residual attention, and produces independent sigmoid-based predictions for each damage category.

## Multi-Scale Feature Extraction Module

MCANet uses Res2Net101 as the feature extraction backbone. Res2Net is built on the residual learning framework but modifies the standard residual block to represent features at multiple spatial scales within a single block (Gao et al. 2021). As shown in Fig. 2, a standard ResNet bottleneck block uses a single $3 \times 3$ convolution between two $1 \times 1$ convolutions. In contrast, the Res2Net block first applies a $1 \times 1$ convolution and then splits the resulting feature channels into multiple subsets. These subsets are processed through a hierarchical split-transform-merge structure, in which later subsets receive information from earlier subsets before convolution. This design creates feature groups with progressively different effective receptive fields and enables multi-scale representation within one residual block.

This intra-block multi-scale design is suited to post-hurricane UAV imagery. Damage-related cues can appear at substantially different spatial extents: debris, vehicles, roof openings, and small pools may occupy localized regions, whereas flooding, blocked roads, and severe structural destruction may require broader scene context. By embedding multi-scale representation within each residual block, Res2Net provides a scale-aware feature tensor for the subsequent class-specific attention module.

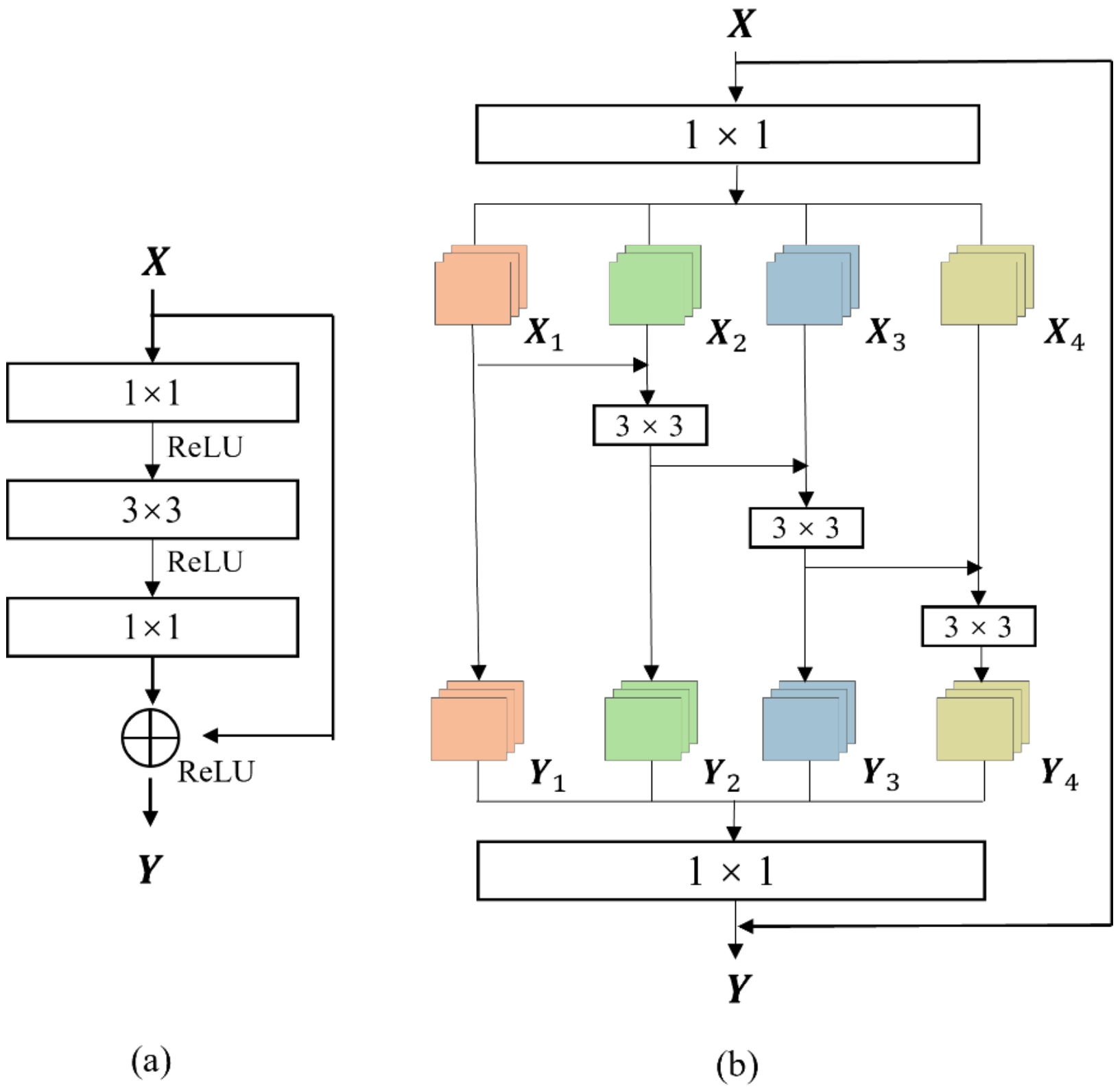

**Fig. 2.** Comparison between (a) a standard ResNet residual block and (b) a Res2Net bottleneck block with intra-block multi-scale feature extraction. The Res2Net block splits feature channels into multiple subsets and processes them through hierarchical connections to capture progressively different spatial scales.

## Class-Specific Residual Attention Module

The Res2Net backbone produces a multi-scale feature tensor $\boldsymbol{X} \in \mathbb{R}^{d \times h \times w}$, where $d$ denotes the number of channels, and $h \times w$ denotes the height and width of the feature map, respectively. For an input image of $448 \times 448$, the final feature tensor has 2048 channels and a spatial resolution of $14 \times 14$. This tensor provides a shared representation for all damage categories before class-specific attention is applied. In multi-label post-hurricane UAV imagery, different categories may correspond to different spatial regions within the same image. For example, a single UAV image may simultaneously contain a blocked road, damaged buildings, vegetation, and water, each requiring attention to different visual evidence. A shared global representation may therefore dilute category-specific cues. To preserve category-dependent spatial information, MCANet uses a class-specific residual attention module (Zhu and Wu 2021).

As shown in Fig. 3, the feature tensor $\boldsymbol{X}$ is flattened into $h \times w$ spatial feature vectors, denoted as $\{\boldsymbol{X}_1, \boldsymbol{X}_2, \ldots, \boldsymbol{X}_{h\times w}\}$, where $\boldsymbol{X}_j \in \mathbb{R}^d$ represents the feature vector at the $j$-th spatial location. For each class $i$, the attention score at spatial location $j$ is computed as:

$$s_j^i = \frac{\exp\left(T \cdot \mathbf{X}_j^T \boldsymbol{m}_i\right)}{\sum_{k=1}^{h\times w} \exp(T \cdot \mathbf{X}_k^T \boldsymbol{m}_i)}, \quad (1)$$

where $\sum_{j=1}^{h\times w} s_j^i = 1$ ensures that the attention scores are normalized across all spatial locations, and $T > 0$ is a temperature parameter controlling the sharpness of the scoring. Here, the vector $m_i \in \mathbb{R}^d$ represents the classifier weights for the $i$-th class. $\mathbf{X}_j^T \boldsymbol{m}_i$ measures the compatibility between the feature vector at the $j$-th spatial location and the classifier weights for class $i$. The temperature parameter $T$ adjusts the focus of attention: larger values of $T$ result in sharper distributions, emphasizing a few key spatial locations, while smaller values produce smoother distributions, spreading the focus across a broader region. Assigning different $T$ values to different attention heads allows the model to capture spatial dependencies at multiple levels of granularity.

The computed attention scores $s_k^i$ are then used as weights to aggregate spatial features into a class-specific feature vector for the i-th class, as follows:

$$\boldsymbol{a}^i = \sum_{k=1}^{\mathrm{h}\times\mathrm{w}} s_k^i \boldsymbol{X}_k. \quad (2)$$

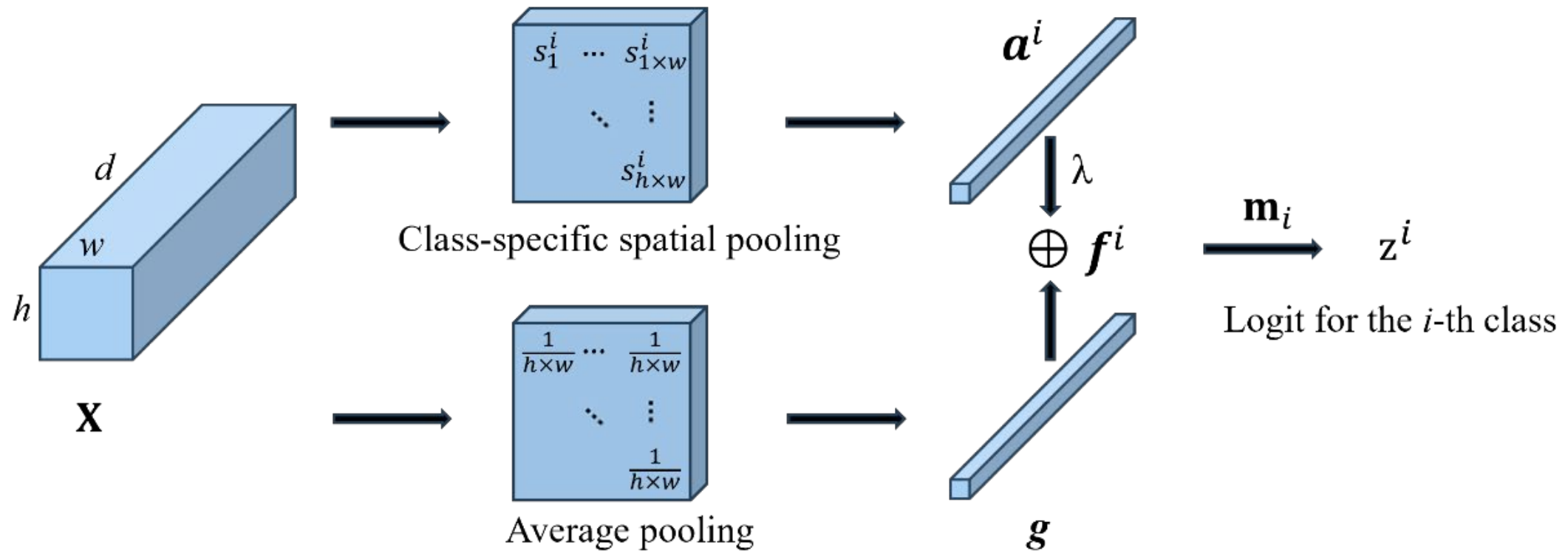

**Fig. 3.** Class-specific residual attention module.

This weighted sum aggregates the spatial features across the image based on their relevance to the $i$-th class, focusing on localized regions of importance for that class. Additionally, a global class-agnostic feature vector $\boldsymbol{g}$ is computed by averaging the feature tensor over all spatial locations:

$$\boldsymbol{g} = \frac{1}{h \times w} \sum_{k=1}^{h \times w} \boldsymbol{X}_k. \tag{3}$$

Finally, as shown in Fig. 3, the global feature vector $\boldsymbol{g}$ and the class-specific feature vector $\boldsymbol{a}^i$ are combined to produce the final class-specific feature vector for the $i$-th class:

$$\boldsymbol{f}^i = \boldsymbol{g} + \lambda \boldsymbol{a}^i. \tag{4}$$

where $\lambda$ is a hyperparameter that controls the relative weight of the global vector $\boldsymbol{g}$ and the class-specific vector $\boldsymbol{a}^i$.

To finalize the classification process, a fully connected layer serves as the classifier, with $\boldsymbol{m}_i \in \mathbb{R}^d$ representing the weight vector corresponding to the $i$-th class. These weight vectors are utilized in conjunction with the class-specific feature vectors $\mathrm{f}^i$ to compute the logits for all classes. The logit for the $i$-th class is calculated as:

$$z^i = \boldsymbol{m}_i^{\mathrm{T}} \boldsymbol{f}^i, \tag{5}$$

where $z^i$ denotes the pre-sigmoid logit for the $i$-th class. The equations above describe the primary single-head CSRA configuration used in MCANet. This study also evaluates a multi-head extension in which multiple CSRA heads are computed using different temperature values and their logits are combined for final prediction. This extension is used to examine whether complementary attention sharpness can improve selected difficult categories.

For the multi-label classification task, the logits are transformed using a sigmoid activation function. The sigmoid function is chosen over the softmax function because it enables independent modeling of each class, a necessity in multi-label classification scenarios where multiple labels can simultaneously apply to a single instance. To obtain the final binary predictions, a threshold of 0.5 is applied to the sigmoid outputs. Output values greater than or equal to 0.5 are classified as 1, indicating the presence of the corresponding class in the image, while values below 0.5 are classified as 0, indicating its absence.

## Loss Function

MCANet was trained with a binary cross-entropy loss computed between the predicted output $\hat{\boldsymbol{y}}_i$ and the actual binary label vector $\boldsymbol{y}_i$. The classes were treated as independent under the multi-label setting, since multiple categories can be present in the same image. Accordingly, the binary cross-entropy loss is computed separately for each class and summed across all classes to obtain the loss for a single instance, which is then averaged over the batch to provide a stable optimization signal during training. The binary cross-entropy loss for a single instance is defined as

$$\text{Loss}(\boldsymbol{y}, \hat{\boldsymbol{y}}) = -\sum_{i=1}^{N} y_i \log\hat{y}_i + (1 - y_i)\log(1 - \hat{y}_i), \quad (6)$$

where $N$ is the total number of classes. The loss is minimized using stochastic gradient descent with momentum.

## Experiments

### Dataset

Hurricane Michael, a Category 5 storm that made landfall near Mexico Beach, Florida, on October 10, 2018, was one of the most intense hurricanes to strike the U.S. mainland, causing over $25 billion in damages and resulting in 16 fatalities (Kennedy et al. 2020). To evaluate MCANet, this study uses the RescueNet dataset, which was collected using a small unmanned aerial system after Hurricane Michael and contains 4,494 red–green–blue (RGB) images for post-disaster scene understanding (Rahnemoonfar et al. 2023). Because the original RescueNet dataset provides pixel-level semantic segmentation masks, the masks were converted to image-level multi-label annotations for the classification task. This produced a binary label vector indicating the damage-related categories present in each UAV image.

As summarized in Table 1, RescueNet includes 10 classes spanning roads, buildings, trees, water, vehicles, and pools. The building category is subdivided into four damage levels: Building No Damage, Building Minor Damage, Building Major Damage, and Building Total Destruction. The road category is divided into Road Clear and Road Blocked. This multi-label structure matches the visual content of post-hurricane UAV imagery, where multiple damage-related categories may co-occur within the same image.

**Table 1.** Damage classification in RescueNet.

| Instance Type | Class Name | Definition |
|---|---|---|
| Road | Road Clear | Roads that are undamaged and clear of water or debris. |
| Road | Road Blocked | Roads that are flooded or obstructed by debris. |
| Tree | Tree | Individual trees or groups of trees. |
| Building | Building No Damage | Buildings with roof damage that are either completely absent or very hardly observable. |
| Building | Building Minor Damage | Buildings with partially damaged roofs that permit the use of tarps as protection. |
| Building | Building Major Damage | Buildings with severely damaged roofs, either entirely or significantly. |
| Building | Building Total Destruction | Fully destroyed buildings with only the base remaining visible. |
| Water | Water | Natural water bodies or floodwater. |
| Vehicle | Vehicle | Cars and trucks. |

| Instance Type | Class Name | Definition |
| --- | --- | --- |
| Pool | Pool | Swimming pools located near buildings. |

Table 2 reports the class label distribution in the combined training and validation set. The distribution is imbalanced across categories. Road Clear is the most frequent class, whereas Road Blocked and Pool are substantially less common. In particular, Pool is the most underrepresented category, appearing in 7.2% of the training and validation images, while Road Blocked appears in 15.4%. This imbalance is relevant to the interpretation of the experimental results because low-frequency classes are generally more difficult to learn and may require more discriminative feature representations. Several representative RescueNet images and their corresponding multi-label annotations are shown in Fig. 4.

**Table 2.** Class label distribution in the combined training and validation set of RescueNet.

| Class name | Count | Percentage of Images (%) |
| --- | --- | --- |
| Road Clear | 2242 | 55.4 |
| Tree | 1895 | 46.9 |
| Building No Damage | 1708 | 42.2 |
| Building Minor Damage | 1529 | 37.8 |
| Vehicle | 1522 | 37.6 |
| Water | 1417 | 35.0 |
| Building Major Damage | 1073 | 26.5 |
| Building Total Destruction | 821 | 20.3 |
| Road Blocked | 624 | 15.4 |
| Pool | 291 | 7.2 |

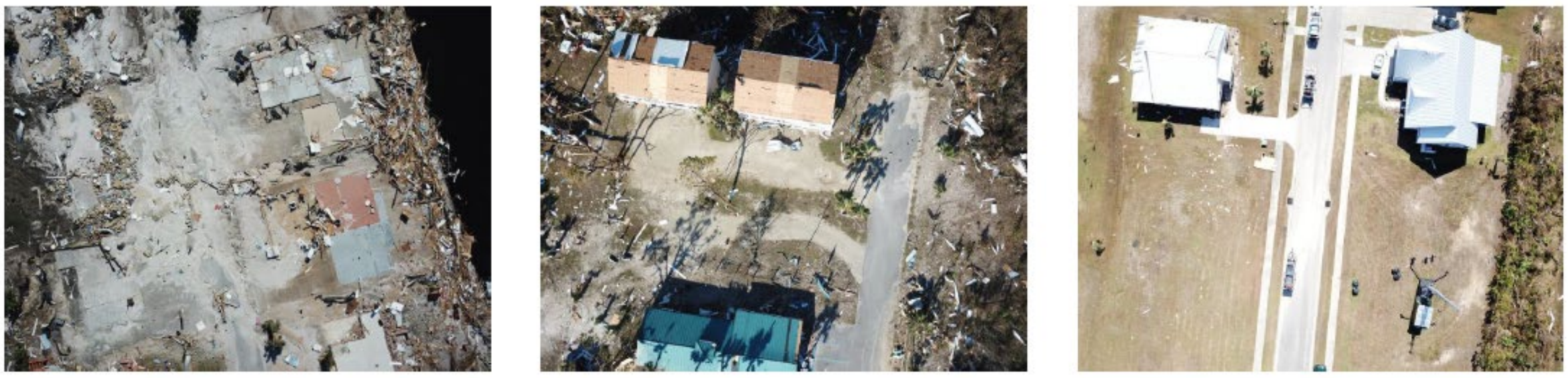

Building Total Destruction, water, Road-Blocked.

Building Minor Damage, Road-Clear.

Building No Damage, Vehicle, Road-Clear, Tree.

**Fig. 4.** Example images and derived multi-label annotations from the RescueNet dataset.

## Experimental Setup

To validate the performance of the proposed MCANet, this paper conducted a series of comparative experiments against several representative architectures. Specifically, the model's effectiveness was evaluated in multi-label hurricane damage classification by benchmarking it against ResNet101 (He et al. 2016), Res2Net101 (Gao et al. 2021), VGG-16 (Simonyan and Zisserman 2015), MobileNetV2 (Howard et al. 2017), EfficientNet-B4 (Tan and Le 2020), and Vision Transformer (ViT-B/16) (Dosovitskiy et al. 2021).
To ensure fair comparison, all baseline models and MCANet were trained under identical experimental conditions: the same train/val/test split (3,595 / 449 / 450 images), input resolution (448 × 448 pixels), data augmentation pipeline (random horizontal flip and random scaled cropping), and ImageNet-pretrained weights for backbone initialization. Hyperparameter settings followed those reported in prior work (Dosovitskiy et al. 2021; He et al. 2016; Zhu and Wu 2021). All models were trained for 30 epochs using stochastic gradient descent with a momentum of 0.9 and a weight decay of $1 \times 10^{-4}$. The learning-rate schedule combined a 2-epoch linear warm-up with a reduce-on-plateau strategy (factor 0.5, patience 5, minimum learning rate $1 \times 10^{-6}$). For CNN-based backbones, the learning rate of 0.01 was applied to pretrained layers and 0.1 to newly initialized components; the same configuration was applied for ViT-B/16. Checkpoints were selected based on best validation mAP.

## Evaluation Metrics

Classification performance was evaluated by comparing predicted labels with the ground-truth labels using four metrics commonly adopted in multi-label image classification: mean average precision (mAP), overall precision (OP), overall recall (OR), and overall F1-measure (OF1), following previous multi-label image classification studies (Wang et al. 2016; Chen et al. 2019; Gao and Zhou 2021).
For class $i$, Average Precision ($\mathrm{AP}_i$) summarizes the precision-recall curve for that class. The mean average precision (mAP) is computed as the macro-average of $AP_i$ across all classes. Overall precision (OP), overall recall (OR), and overall F1 (OF1) are computed by aggregating true positives, false positives, and false negatives across all classes and then applying the standard precision, recall, and F1 definitions. mAP and OF1 are used as the primary evaluation metrics. The evaluation metrics are defined as follows:

$$\mathrm{AP}_i = \int_0^1 P_i(R)dR \tag{7}$$

$$\mathrm{mAP} = \frac{1}{N}\sum_{i=1}^{N} AP_i \tag{8}$$

$$\mathrm{OP} = \frac{\sum_i N_i^c}{\sum_i N_i^p} \tag{9}$$

$$\mathrm{OR} = \frac{\sum_i N_i^c}{\sum_i N_i^g} \tag{10}$$

$$\mathrm{OF1} = \frac{2 \times \mathrm{OP} \times \mathrm{OR}}{\mathrm{OP} + \mathrm{OR}} \tag{11}$$

where $P_i(R)$ is the precision as a function of recall R, $N$ is the total number of classes, $AP_i$ represents the average precision for the $i$-th class. $N_i^c$ is the number of images correctly predicted for the i-th class, $N_i^p$ is the number of images predicted as positive for the i-th class, $N_i^g$ is the number of ground truth images for the i-th class. For each image, a label is predicted as positive if its estimated probability exceeds the threshold of 0.5. Throughout the paper, all reported differences in mAP, AP, OP, OR, and OF1 are absolute differences in percentage points.

## Results and Analysis

### Overall Comparison with Baseline Models

Table 3 summarizes the performance of MCANet and the baseline models on the RescueNet test set. MCANet achieves the highest mAP at 91.37%, outperforming all compared baselines in this primary metric. It also attains the highest overall recall (87.22%) and OF1 (87.34%), while ResNet101 achieves the highest overall precision. MobileNetV2 and ViT-B/16 remain competitive, with mAP values of 89.54% and 89.47%, respectively, whereas VGG-16 and EfficientNet-B4 show lower overall performance. Overall, these results indicate that MCANet provides a favorable accuracy-recall balance on the RescueNet test set.

Compared with Res2Net101, MCANet improves mAP by 0.69 percentage points. It also exceeds ResNet101 by 0.87 percentage points. Notably, MCANet outperforms ViT-B/16 by 1.90 percentage points in mAP while requiring substantially fewer parameters and giga floating-point operations (GFLOPs), indicating a more favorable accuracy-efficiency trade-off. The class-level results further show that these gains are not uniform across all categories but are concentrated in several difficult classes, as analyzed in the following subsection.

**Table 3.** Performance comparison of different methods in multi-label post-hurricane damage classification.

| Method | mAP | OP | OR | OF1 |
|---|---|---|---|---|
| ResNet101 | 90.50 | **87.88** | 85.07 | 86.45 |
| Res2Net101 | 90.68 | 86.70 | 85.85 | 86.28 |
| VGG-16 | 83.68 | 81.78 | 79.86 | 80.80 |
| MobileNetV2 | 89.54 | 86.99 | 84.09 | 85.52 |
| EfficientNet-B4 | 86.91 | 84.18 | 76.66 | 80.25 |
| ViT-B/16 | 89.47 | 86.20 | 84.68 | 85.43 |
| **MCANet** | **91.37** | 87.45 | **87.22** | **87.34** |

Note: Bold values indicate the best result in each column.

### Per-Class Performance Analysis

Per-class results provide further insight into how MCANet behaves across different damage categories. As shown in Table 4, classes with stable and visually distinctive patterns, such as Water, Tree, and Road Clear, achieve high AP values across most models. By contrast, three groups of categories remained more difficult: those that depend on subtle structural differences (Building Minor Damage, Building Major Damage), those that center on small localized objects (Pool, Vehicle), and those that depend on cluttered scene-level context (Road Blocked). In the per-class comparison, MCANet achieves the highest AP in Road Blocked and Pool and remains competitive in several additional categories, including Building No Damage, Building Minor Damage, and Vehicle.

**Table 4.** Per-class average precision (%) of all evaluated methods on the RescueNet test set.

| Category | VGG-16 | EfficientNet-B4 | ViT-B/16 | MobileNetV2 | ResNet101 | Res2Net101 | MCANet |
|---|---|---|---|---|---|---|---|
| Water | 92.56 | 92.34 | 96.20 | 93.69 | **96.70** | 95.93 | 95.80 |
| Building No Damage | 92.66 | 91.36 | 93.20 | **94.91** | 93.68 | 93.49 | 94.61 |
| Building Minor Damage | 82.83 | 82.96 | **89.15** | 85.64 | 87.36 | 84.71 | 88.85 |
| Building Major Damage | 76.16 | 76.92 | 85.00 | 83.15 | **86.32** | 86.23 | 85.94 |
| Building Total Destruction | 80.15 | 86.01 | 87.42 | 88.08 | **91.21** | 90.71 | 88.88 |
| Vehicle | 91.52 | 92.42 | 93.37 | 94.02 | **95.09** | 93.67 | 94.85 |
| Road Clear | 94.27 | 96.75 | 98.32 | 97.42 | 98.07 | **98.36** | 98.15 |
| Road Blocked | 55.92 | 63.86 | 68.83 | 69.59 | 68.00 | 71.58 | **72.90** |
| Tree | 91.64 | 93.73 | **96.69** | 96.07 | 96.64 | 95.82 | 96.00 |
| Pool | 79.10 | 92.71 | 86.51 | 92.79 | 91.91 | 96.25 | **97.72** |

Note: Bold values indicate the best result in each row.

The building-related categories are particularly challenging because they form a visual continuum from no damage to total destruction. Building Minor Damage is especially important in this context because it depends on subtle evidence that can be difficult to separate from both undamaged and more severely damaged structures. MCANet achieves an AP of 88.85% for this category, which is higher than both ResNet101 and Res2Net101 and close to the strongest baseline overall. Road Blocked and Pool further illustrate the strengths of the proposed model. Road Blocked is one of the most difficult categories across all evaluated models, reflecting both its lower frequency and its substantial visual variability. This category is also practically important because its recognition can affect downstream road accessibility assessment and emergency routing. MCANet achieves the highest AP for this class at 72.90%, which is 4.90 percentage points higher than ResNet101. Pool, although also underrepresented, reaches 97.72% AP under MCANet, again the highest among all evaluated models.

Taken together, these results suggest that the value of MCANet lies less in uniform gains across all categories than in improved recognition of selected difficult classes that are more ambiguous and underrepresented in post-disaster assessment. These class-level patterns motivate a closer examination of the contributions of the backbone and attention design.

## Component Ablation Study

To examine the contributions of the multi-scale backbone and class-specific attention design, a component ablation study was conducted using ResNet101 without attention as the reference configuration. As shown in Table 5, replacing ResNet101 with Res2Net101 increases mAP from 90.50% to 90.68%, indicating a modest aggregate improvement from the multi-scale backbone. However, the per-class results provide a more informative interpretation of this component. Compared with ResNet101, Res2Net101 without attention improves AP for Road Blocked by 3.58 percentage points and for Pool by 4.34 percentage points. This suggests that the backbone contribution is concentrated in scale-sensitive categories, consistent with the multi-scale rationale of the Res2Net block.

**Table 5.** Component ablation results on the RescueNet test set.

| Method | mAP | OP | OR | OF1 | ΔmAP vs. ResNet101 |
|---|---|---|---|---|---|
| ResNet101 | 90.50 | **87.88** | 85.07 | 86.45 | — |
| ResNet101 + CSRA | 91.19 | 87.24 | **87.35** | 87.30 | +0.69 |
| Res2Net101 | 90.68 | 86.70 | 85.85 | 86.28 | +0.18 |
| Res2Net101 + CA | 89.85 | 87.81 | 84.55 | 86.15 | -0.65 |
| **MCANet** | **91.37** | 87.45 | 87.22 | **87.34** | **+0.87** |

Note: Bold values indicate the best result in each column.

The attention ablation further clarifies that the benefit of MCANet does not come from spatial attention alone, but from class-specific spatial modeling. When class-agnostic attention (CA) is applied to the Res2Net101 backbone, mAP decreases from 90.68% to 89.85%. In contrast, replacing the shared attention map with class-specific residual attention increases mAP to 91.37%, which is 1.52 percentage points higher than class-agnostic attention and 0.69 percentage points higher than the attention-free Res2Net101 baseline. This comparison suggests that a single shared attention map may be dominated by visually prominent regions, thereby suppressing evidence for less frequent or spatially localized classes. By computing attention separately for each category, CSRA better preserves class-specific evidence and better matches the multi-label structure of post-hurricane scenes.

The benefit of CSRA is also observed across different backbone settings. CSRA also improves the ResNet101 backbone (90.50% → 91.19%), suggesting that the gain from class-specific spatial attention is not specific to Res2Net101. Combined, the multi-scale backbone and CSRA raised OR from 85.07% to 87.22%, and OF1 from 86.45% to 87.34% relative to the ResNet101 baseline. These ablation results identify class-specific residual attention as the main source of the aggregate mAP improvement, while the Res2Net backbone contributes more selectively at the class level.

## Multi-Head Attention Analysis

To examine whether multiple attention heads provide additional benefit, MCANet was evaluated under different CSRA head configurations. As shown in Table 6, the single-head setting achieves the highest overall mAP and is therefore used as the default MCANet configuration. The two-head setting provides targeted gains for selected difficult categories: Road Blocked AP increases from 72.90% to 76.77%, and Building Major Damage AP increases from 85.94% to 89.11%. These results suggest that a limited multi-head design can improve categories that require both localized evidence and broader spatial context. However, the three-head and four-head settings do not further improve these selected categories or surpass the single-head configuration in overall mAP. Thus, increasing the number of attention heads introduces class-level trade-offs rather than uniform aggregate gains.

**Table 6.** Effect of CSRA head count on overall performance and selected difficult categories.

| Heads | mAP (%) | Road Blocked AP (%) | Building Major Damage AP (%) |
|---|---|---|---|
| 1 | **91.37** | 72.90 | 85.94 |
| 2 | 91.20 | **76.77** | **89.11** |
| 3 | 91.17 | 71.30 | 87.40 |
| 4 | 90.87 | 69.51 | 87.32 |

Note: Bold values indicate the best result in each column.

## Model Interpretability via Class Activation Mapping

To examine the spatial evidence used by different methods, class activation maps (CAMs) (Zhou et al. 2016) were visualized for a representative RescueNet test image. For MCANet, the maps were obtained from the class-specific responses in the CSRA module, whereas for Res2Net101 and ResNet101, CAMs were generated from the final convolutional features and classification weights using the standard CAM formulation. All maps were resized to the input image resolution for visualization.

Fig. 5 compares the activation maps of MCANet, Res2Net101, and ResNet101 for five selected classes. For this image, the remaining five classes are ground-truth negative and are correctly suppressed by all three models. The displayed classes were selected to include both disagreement cases and correctly detected categories, allowing comparison of spatial localization and class discrimination behavior. The clearest differences appear in the two adjacent building-damage categories. For Building Minor Damage, which is absent in this image, MCANet assigns a low probability and suppresses activation over the severely damaged structures, while Res2Net101 produces a strong false positive response. For Building Major Damage, MCANet correctly identifies the class and concentrates its activation on the relevant damaged structures, whereas the two baseline models fail to produce meaningful responses. By contrast, for Building Total Destruction, Vehicle, and Road Clear, all three models generate correct predictions and reasonable activation patterns, although MCANet remains visually more focused in some regions. These activation patterns are consistent with the per-class AP differences in Table 4, particularly for the adjacent building-damage categories.

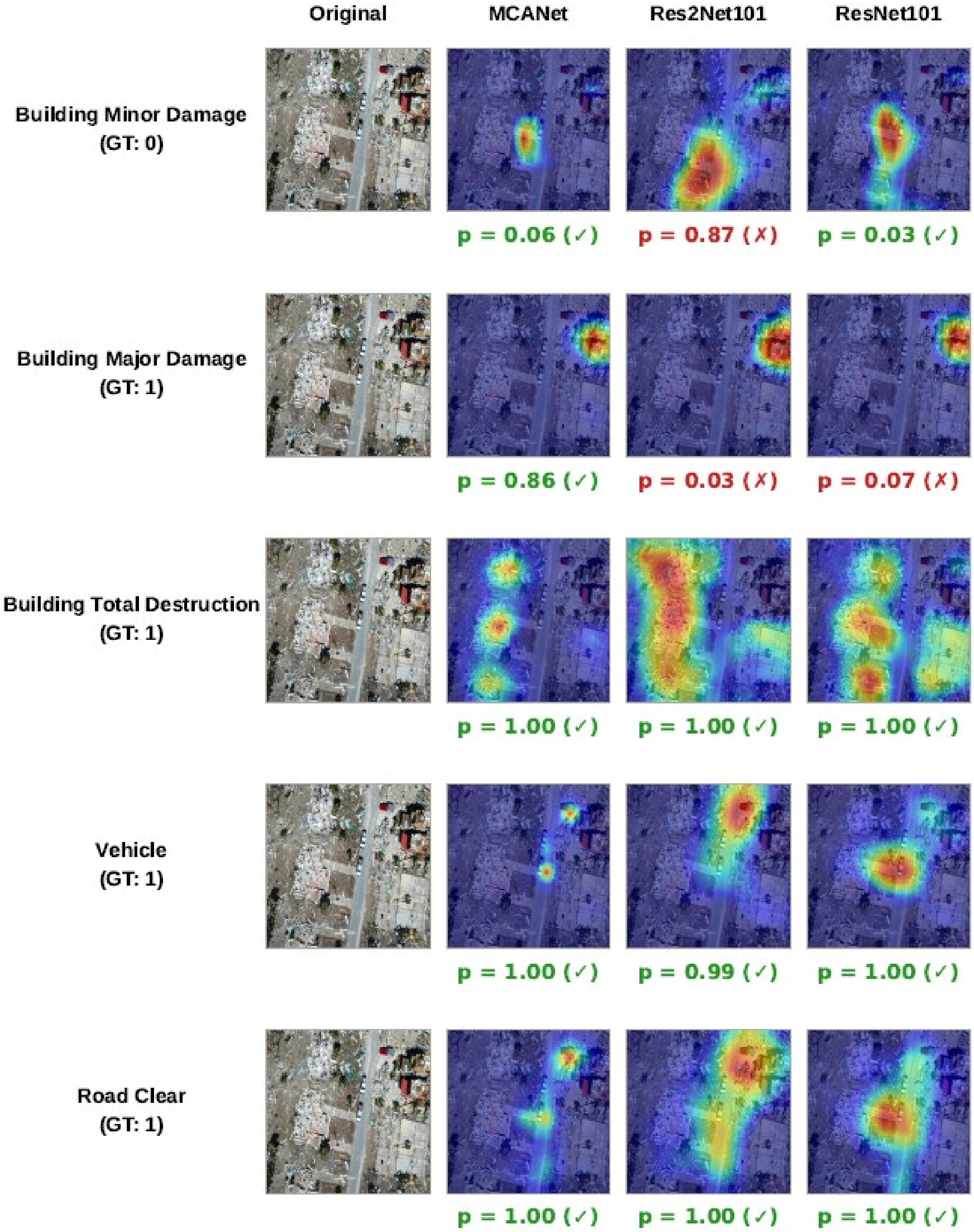

**Fig. 5.** Comparison of class activation maps for MCANet, Res2Net101, and ResNet101 on a representative RescueNet test image. The selected classes illustrate differences in spatial localization and discrimination, particularly for visually similar building-damage categories.

## Computational Analysis

MCANet's computational cost was compared with the baselines in terms of parameter count, GFLOPs, per-epoch training time, and inference latency. All measurements were obtained on an NVIDIA A100 Peripheral Component Interconnect Express 40 GB graphics processing unit (GPU) using 448 × 448 input images, batch size 1 for inference, and a 500-run Compute Unified Device Architecture synchronized average. Table 7 summarizes the results.

**Table 7.** Computational comparison of MCANet and baseline models on the RescueNet test set.

| Model | Params (million) | GFLOPs | Train (min/epoch) | Latency (ms) |
|---|---|---|---|---|
| MobileNetV2 | 2.24 | 1.30 | 2.0 | 5.82 |
| EfficientNet-B4 | 17.57 | 6.30 | 2.1 | 17.24 |
| ResNet101 | 42.52 | 31.46 | 2.1 | 14.82 |
| ViT-B/16 | 86.26 | 67.20 | 3.8 | 25.05 |
| Res2Net101 | 43.18 | 32.56 | 2.2 | 28.87 |
| MCANet | 43.18 | 32.56 | 2.2 | 28.82 |
| VGG-16 | 134.30 | 61.51 | 2.1 | 4.49 |

MCANet $(h = 1)$ contains 43.18M parameters and requires 32.56 GFLOPs per forward pass. Compared with ViT-B/16, it uses approximately 50% fewer parameters and 52% fewer GFLOPs, while achieving a higher mAP (91.37% versus 89.47%). This result indicates that MCANet provides a more favorable accuracy-efficiency trade-off than the transformer baseline on RescueNet. By contrast, VGG-16 has the largest parameter count yet the lowest mAP, suggesting that larger model size alone does not translate into better performance for this task. The additional computational cost of the CSRA module is negligible. Compared with the attention-free Res2Net101 baseline, MCANet achieves nearly identical inference latency (28.82 ms vs. 28.87 ms) and the same per-epoch training time (2.2 min), while improving mAP from 90.68% to 91.37%.

With an inference latency of 28.82 ms per image, MCANet can process approximately 34.7 images per second on a single A100 GPU. At this throughput, a post-hurricane UAV survey containing 5,000 to 10,000 images could be analyzed in approximately 2.4 to 4.8 minutes. MCANet also maintains modest memory demand, with a measured peak memory of 1591.3 MB; relative to the attention-free Res2Net101 baseline, the CSRA module adds only 179.8 MB. These results show that MCANet is computationally feasible for rapid post-disaster image screening in server-side or cloud-assisted workflows. Although lightweight models such as MobileNetV2 offer lower latency, they do so at the cost of reduced classification accuracy, which may be less suitable when missed damage categories could affect operational decisions.

# Discussion

This study contributes a multi-label damage classification framework that connects UAV-based visual recognition with the information needs of post-hurricane response. The multi-label formulation is important because post-disaster UAV images often contain multiple co-occurring scene conditions, such as building damage, road blockage, vehicles, water, trees, and pools. Compared with single-label classification, this formulation better reflects the mixed visual content of real disaster scenes (Qi et al. 2020; Shao et al. 2018). Compared with pixel-level segmentation, the multi-label formulation offers a more lightweight image-level screening approach by relying on image-level labels rather than dense pixel-level outputs. This is practical for rapid post-disaster data processing, particularly when the initial response need is to identify which damage-related conditions are present before conducting more detailed localization (Rahnemoonfar et al. 2023).

A second contribution is that MCANet links architectural design choices to specific recognition challenges in post-hurricane UAV imagery. The Res2Net backbone provides a multi-scale feature representation for damage cues that vary in spatial extent. Although the aggregate mAP gain from the backbone alone is limited, multi-scale representations benefit categories that combine localized cues with broader scene context, consistent with the per-class gains observed for Road Blocked and Pool. The ablation results further show that the value of attention in this setting depends on preserving class-specific spatial evidence. Class-agnostic attention performs worse than the attention-free Res2Net101 baseline, whereas class-specific residual attention improves mAP over both. This pattern is consistent with the structure of post-hurricane imagery, where shared attention may emphasize dominant visual evidence while CSRA preserves category-specific responses for coexisting labels. The head-count analysis indicates a trade-off rather than a uniform gain: additional heads can benefit selected difficult categories, whereas the single-head configuration remains the default setting in overall mAP.

The experimental results show that MCANet improves overall classification performance while providing clearer benefits for selected difficult categories that are visually ambiguous, underrepresented, or operationally important. MCANet achieves the highest AP for Road Blocked and Pool and remains competitive for Building Minor Damage. These class-level patterns are meaningful for disaster response. Road Blocked is directly related to road accessibility and emergency routing, while building-damage severity affects inspection prioritization and recovery planning. The class activation map visualizations provide qualitative support for this interpretation by showing that MCANet can suppress irrelevant activation for absent damage classes and focus more consistently on relevant damaged structures. This qualitative evidence is consistent with the quantitative results and suggests that class-specific attention improves the interpretability of model decisions in visually complex scenes.

From an operational perspective, MCANet is suitable for server-side or cloud-assisted post-disaster image screening. UAVs can collect high-resolution imagery over affected areas, and MCANet can convert these images into structured multi-label outputs. When combined with UAV geolocation data, these outputs can support multi-layered damage maps that include building damage severity, road accessibility, water presence, and other scene-level conditions. Such information can provide input to downstream emergency management workflows, including search-and-rescue prioritization, routing support (Pan et al. 2024), UAV flight path optimization (Hu et al. 2023), and digital twin–enabled situational awareness (Mohammadi and Taylor 2017, 2021). The computational results support the feasibility of MCANet for post-disaster image triage, where survey-scale UAV image batches can be processed within a short operational window. The small marginal cost of CSRA over the attention-free backbone suggests that the accuracy gains do not require substantial additional deployment overhead. This trade-off is important for disaster response, where faster lightweight models may miss damage categories that affect downstream assessment and prioritization. Onboard UAV deployment would require further evaluation under memory, power, and hardware constraints.

This study also presents several limitations. First, this study is evaluated on the RescueNet dataset, which provides a valuable post-hurricane UAV benchmark but represents a single disaster context. Additional validation across other hurricanes, geographic regions, and sensing conditions is needed to assess external generalizability. Second, class imbalance remains a challenge. Low-frequency categories such as Road Blocked and Pool are important for response planning but appear less frequently in the dataset, which can affect model reliability. Future work may explore targeted data augmentation, focal or asymmetric losses, and synthetic image generation to improve performance on underrepresented classes (Lu et al. 2025). Third, the current model relies on RGB UAV imagery and may be affected by poor illumination, occlusion, motion blur, or low-resolution imagery. Integrating additional modalities, such as infrared imagery, geospatial layers, social media reports, or environmental sensor data, could improve robustness under more diverse disaster conditions (Huang et al. 2025). Future research could also extend MCANet beyond image-level classification by linking its outputs with disaster-specific knowledge graphs and multimodal foundation models. Knowledge graphs could encode relationships among damage types, infrastructure elements, and environmental features (Liu et al. 2025c; Thomas et al. 2025), while multimodal large language models could support richer semantic interpretation, improve generalization to unseen damage categories, and enhance decision support (Ho et al. 2025; Huang et al. 2026).

## Conclusion

This study presents MCANet, a multi-label classification framework for post-hurricane damage assessment using UAV imagery. MCANet combines a Res2Net multi-scale backbone with class-specific residual attention for multi-label damage classification in post-hurricane UAV imagery. This design allows the model to represent co-occurring damage categories while preserving category-specific spatial evidence. Evaluation on the RescueNet dataset shows that MCANet achieves the highest mAP of 91.37% among the evaluated models, outperforming ResNet101, Res2Net101, VGG-16, MobileNetV2, EfficientNet-B4, and ViT-B/16. Beyond the overall mAP gain, MCANet improves recognition of difficult categories, particularly Road Blocked and Pool. The ablation study shows that class-specific attention provides a clear benefit over class-agnostic attention. Computational analysis further indicates that MCANet offers a favorable accuracy-efficiency trade-off. Overall, MCANet provides a practical and interpretable framework for multi-label UAV-based hurricane damage assessment and its structured multi-label outputs can support downstream emergency-management workflows. Future work should evaluate the framework across additional disaster events and sensing conditions, and explore integration with multimodal data sources, disaster-specific knowledge graphs, and foundation models to improve adaptability and decision-support capability.

## Data Availability Statement

Some or all data, models, or code generated or used during the study are available in a repository online in accordance with funder data retention policies. The RescueNet dataset used in this study is available as described by (Rahnemoonfar et al. 2023), and the MCANet code is available through GitHub (Liu 2025d).